\documentclass[10pt,twocolumn,letterpaper]{article}

\usepackage{graphicx}
\usepackage{amsmath}
\usepackage{amssymb}
\usepackage{booktabs}
\usepackage{multirow}

\usepackage[pagebackref,breaklinks,colorlinks]{hyperref}

\usepackage[capitalize]{cleveref}
\crefname{section}{Sec.}{Secs.}
\Crefname{section}{Section}{Sections}
\Crefname{table}{Table}{Tables}
\crefname{table}{Tab.}{Tabs.}

\def\cvprPaperID{7853} 
\def\confName{CVPR}
\def\confYear{2022}

\begin{document}

\title{Extracting knowledge from features with multilevel abstraction}

\author{
{Jinhong Lin \space Zhaoyang Li} \\
University of Wisconsin Madison\\
{\tt\small jlin398@wisc.edu \space \tt\small zli2344@wisc.edu}}


\maketitle

\begin{abstract}
   Knowledge distillation aims at transferring the knowledge from a large teacher model to a small student model with great improvements of the performance of the student model. Therefore, the student network can replace the teacher network to deploy on low-resource devices since the higher performance, lower number of parameters and shorter inference time. Self-knowledge distillation (SKD) attracts a great attention recently that a student model itself is a teacher model distilling knowledge from. To the best of our knowledge, self knowledge distillation can be divided into two main streams: data augmentation and refined knowledge auxiliary. In this paper, we purpose a novel SKD method in a different way from the main stream methods. Our method distills knowledge from multilevel abstraction features. Experiments and ablation studies show its great effectiveness and generalization on various kinds of tasks with various kinds of model structures. Our codes have been released on GitHub.
\end{abstract}

\section{Introduction}
\label{sec:intro}

Since Deep neural networks (DNNs) were proposed, there have been many remarkable successes in various fields of computer vision but the success of DNNs often depends on the computing and storage capabilities that may be restricted on mobile devices. To mitigate the problem, knowledge distillation(KD) was proposed\cite{hinton2015distilling} to transfer knowledge from a large model(teacher model) to a small model(student model) and deploy the small one on edge devices to achieve great performance and fast inference. Although KD can save computational resources and inference time on edge devices, pretraining the large model still causes substantial resource burdens. To reduce the need for such large models, some methods were proposed, such as deep mutual learning(DML)\cite{zhang2018deep}. For DML, multiple networks are used for training simultaneously and these work improve the generalization ability by referring each other.Although using DML can reduce the necessary to train a large teacher network, there is still the computational resource burden since multiple models are trained together. Furthermore, recent studies focus on self-knowledge distillation(self-KD)\cite{wang2020eventsr, zhang2019your}, which progressively improve the performance of the network, does not require pretrained teacher network.
 
  Self-distillation can be achieved by three methods: transferring knowledge based on feature maps and that based on soft label. Studies, such as \cite{hou2019learning}, achieve great performance via transferring knowledge from refined feature maps to grained ones, but such approach causes relatively high computation load. On the other hand, transferring knowledge with soft label \cite{ji2021refine} is convenient and low computation burdens to train shallow layers using dark knowledge from deep layers. And the third way is to use both of above two methods. Such as \cite{zhang2019your, ji2021refine}, where refined knowledge extracted from deeper layers is used to guide shallow layer via feature maps and soft label. This method usually can achieve great performance but bring higher computation burden too. In this paper, we propose a novel self-distillation method, Extracting  Knowledge from  Features  with  Multilevel  Abstraction(LFMA), is to transfer knowledge based on soft label, which is a low computation burden method.


  No matter teacher-student\cite{hinton2015distilling} or self-distillation\cite{wang2020eventsr} \cite{zhang2019your}, knowledge is transferred from the refined part to the coarse part. In the teacher-student knowledge distillation\cite{hinton2015distilling}, knowledge from the teacher model is more refined and can bring a higher accuracy; in self-distillation\cite{zhang2019your}, knowledge from the deeper layer is used to guide the sallow layer through soft label and feature map. Our work is focused on the opposite direction which is using coarse-grained knowledge to train deep layers auxiliary. Coarse-grained knowledge extracted from the shallow layers is much general, like object edges, eyes. Fine-grained knowledge extracted from deep layers are much abstract and specific, like facial features\cite{zintgraf2017visualizing}. Different grained knowledge reveals different dark knowledge\cite{hinton2015distilling} in different aspects, and which provides enough details and relationship\cite{hinton2015distilling} to train networks for improving better performance. 
  
  In our method, LFMA, multilevel abstraction features extracted from different layers is used to guide the deepest output and achieve state-of-the-art performance in image classification task on various datasets with soft label guided self-distillation. And it provides a novel perspective to distillation method and dark knowledge.
  
  In summary, we make the following principle contributions in this paper:
  \begin{itemize}
      \item Our method, LFMA, achieves great performances among soft label guided self-distillation methods on various datasets.
      \item We put forward an augmentation based on feature maps, which can achieve high efficiency data augmentation. 
      \item We provide a novel perspective to distillation method and dark knowledge.
      \item Experiments on four different scale datasets to prove the generalization of our method.
  \end{itemize}

  

\begin{figure*}
	\centering
    \includegraphics[width=14cm]{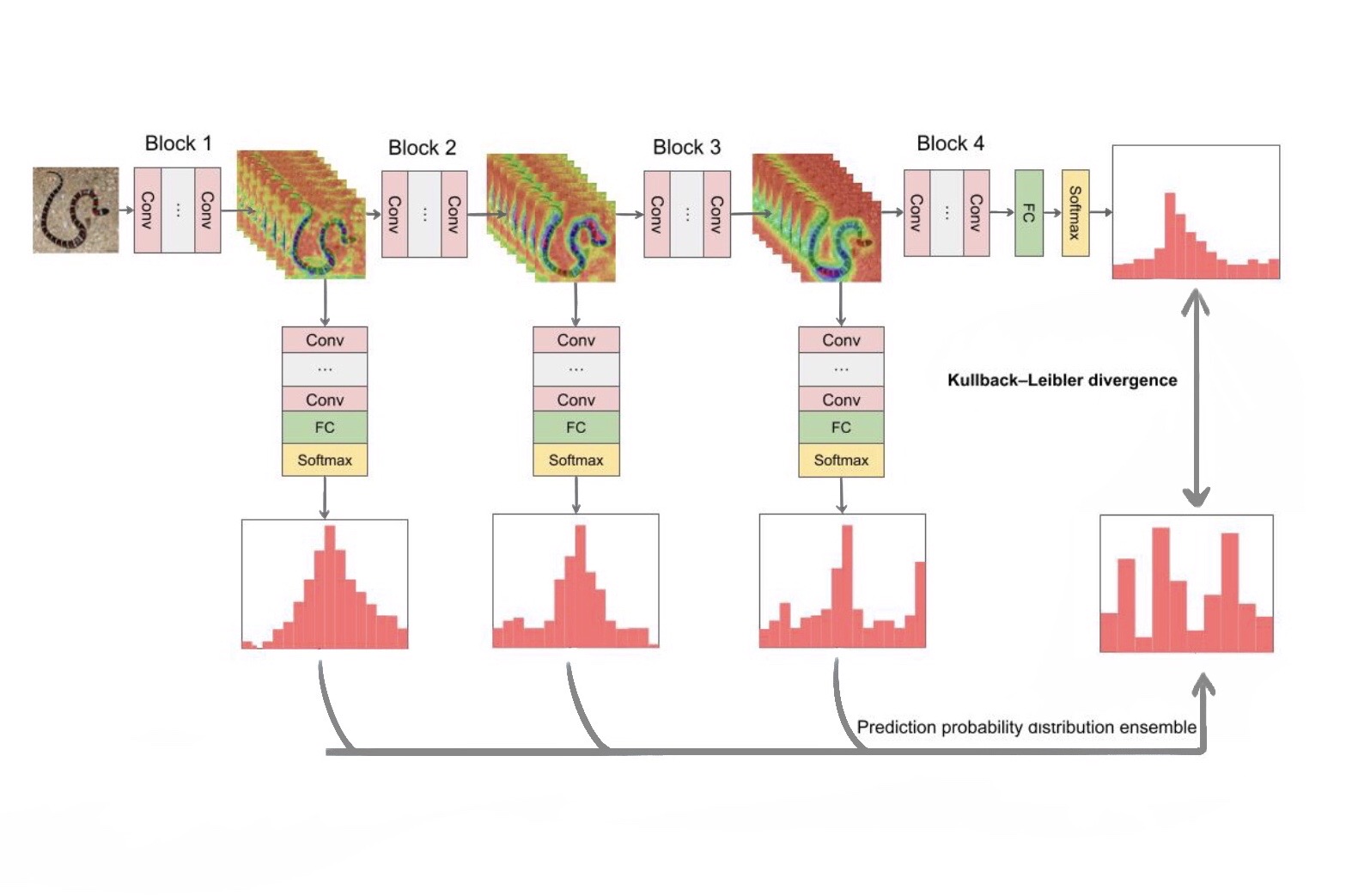}
    \caption{Overview of our method, LFMA. Given a input image,  the blocks output various level abstraction feature maps. 
    They are fed into extra layers to process and used to predict probability distributions, which are ensemble into a mixed distribution by weighted average. The ensemble distribution and the ground true labels from datasets are used to guide outputs of every blocks.
    }
    \label{fig:Overview}    
\end{figure*}

\section{Related Work}

 \textbf{Knowledge distillation(KD):} KD\cite{hinton2015distilling} is a technique used to compress a teacher network to a student network while maintaining the performance. The student model learns from ground true labels of the dataset and soft labels from the teacher model; using a higher temperature to shift the soft probability distribution enables a more useful soft label, and this a one method of KD from logits. And then, Hinton \cite{hou2019learning} proposed to perform knowledge distillation by minimizing the distance between student and teacher network output distribution statistics. Furthermore, Fitnets first introduced hints, extracted feature results from the middle layer of the teacher network to train student networks \cite{romero2014fitnets}. Further utilizing Fintnets, the flow of solution procedure (FSP) that fits the relationship between layers of the large model is also used to transfer knowledge between networks \cite{yim2017gift}. And Zhou et al \cite{zhou2020channel}. proposed each channel of feature maps corresponds with a visual model, so they focus on transferring attention concepts of feature map \cite{wu2018image, wu2019pseudo} from each channel of intermediate layers However, some exiting approaches do not make better of different knowledge from different layers. Form each intermediate layer, the level of knowledge is different. The intermediate knowledge has large potential uses. In our paper, we will make more effective use of the dark matter in the middle layer so that the network can also use the dark matter that the model may not use.
    
  \textbf{Self-distillation: }Self-distillation means that the model distils itself, and we can say that the teacher model is an integrated version of the student model. There some approaches that introduce an informative thought in self-distillation. BYOT \cite{zhang2019your} suggested that the last block of networks is teacher and the rest of the shallow blocks are students, where feature maps and soft labels generated from the teacher are used to train the student. Based on the training teacher network, it uses KD to train student networks. Furthermore the paper, the paper \cite{kim2020self} posed the current model as a student and the previous model as teacher perform KD to train . Self-distillation greatly increases the accuracy of a model after training and shorten the training time of the model. Due to the advantages of self-distillation, we take this model instead of others. Self-distillation can accelerate the efficiency of training and easily extract knowledge from intermediate layers. Based on KD, the paper\cite{ji2021refine} generate refine feature maps and soft label for training network itself. In our paper, we use a different way than the way which use relatively refine knowledge to train the network, we use different multi-features from different network layers to provide rich dark knowledge to help network training.
  
  \textbf{Deep Supervision: }Deeply-supervised nets (DSN) utilizes classifiers at hidden layers to minimize classification errors and improve the performance inference. DSN adds the extra layer and trains this extended layer, supervising the whole networks we are training \cite{blum1998combining}. And this approach is used to solve the problems of DNNs training gradient and slow convergence speed. Applications about DSN are widely used, such as image classification and medical images segmentation and so forth \cite{blum1998combining,dou20163d,yu2017volumetric}. This method can effectively combine with other models. For example, in the paper Deeply-supervised Knowledge Synergy effectively combine with distillation; it is an improved algorithm for knowledge distillation in the middle layer\cite{sun2019deeply}. Better generalization ability is obtained by adding additional supervision branches in some intermediate layers. In addition, a novel cooperative (snergy) loss function is proposed to consider the knowledge matching between all supervised branches through the distance between features. Therefore, we utilize deep supervision to interact with self-distillation and strengthen the model we create.

  \textbf{Multi-view Data: }In the real world, a data object often has multiple attribute sets at the same time; every attribute set constitutes a view. And there are often some views in a model, and we can utilize multiple views to improve our training. One of the co-training methods is a typical example of training Multi-view data\cite{sun2019deeply}. There are some papers talk about utilizing multi-view. For example, one of paper considers that almost all perspective features will be displayed. However, in the input pixel space and in the middle layer perspective features exit\cite{allen2020towards}. That is to say, we can extract multiple views in intermediate layers, and these views have different knowledge levels, which represent the thickness of knowledge. We can get more practical knowledge from multiple-view data.

\section{Methodology}
This section introduces our proposed self-distillation method, Features  with  Multilevel  Abstraction(LFMA), in detail. 

Given samples $\{x_i, y_i\}_{i=1}^{N}$ in dataset from the dataset, where $i$ denotes the index of the sample in the dataset, $N$ denotes the number of the dataset, $x_i$ and $y_i$ correspond to the image and its label in the dataset for the image classification task.
We introduce Extra Layers to process feature maps from middle blocks of the backbone network to get prediction probability distribution. And those distributions are ensemble by some ways(e.g., average) to an ensemble probability distribution.Features are augmented with Feature Map Cutout techniques before passing into Extra Layers in order to improve the generalization of models. The Kullback–Leibler divergence loss is applied to compute the ensemble probability distribution with the output of the backbone. The Extra Layers are applied in training and can be removed in inference, so our method does not increase any parameters and decrease inference time in practice.

We introduce Motivation, Extra Layer and Ensemble Probability Distribution and Feature Map Cutout in details.

\subsection{Motivation}
 Previous self-knowledge distillation methods tend to exploit the most abstract features to guide the shallow layers for training. Inspired by Feature Pyramid Networks\cite{lin2017feature}, we realize that multilevel abstraction features used together can be more suitable to play the role of teacher. The technique ensembles features with different abstraction to predict is efficient in object detection\cite{redmon2018yolov3} since those features provide different scale information of images: the high abstract features extracted from deep layers are object shape and etc., like human face\cite{zintgraf2017visualizing}; the low abstract features extracted from the shallow layers are edges of objects and etc.\cite{zintgraf2017visualizing}. So we ensemble features with multilevel abstraction generated from different layers to auxiliary training.


\subsection{Extra Layer}
Since feature maps from middle blocks are specific to the final layer in the backbone, they cannot be used to predict directly. Therefore, we introduction Extra Layers to process such feature maps and make them suitable for prediction. 

Extra layers are elastic. We can take different parameters and structures networks as Extra layers. They can be formulated as:
\begin{equation}
    D_{B_i} = F_i(f_{B_i})
\end{equation}

where $i$ denotes the index of block in the backbone, $B_i$ denotes the ith block, $f_{B_i}$ denotes the feature maps generated by the $B_i$, $F_i$ denotes the Extra layers, FC layer and the softmax function corresponding to the ith block, and $D_{B_i}$ denotes the prediction probability distribution of the output of $F_i$.

	

 \subsection{Ensemble Probability Distribution}

Probability distributions reveal relationships of different classes in certain aspects, and they are depicted as dark knowledge\cite{hinton2015distilling}. Those distributions based on coarse features represent the similarities of classes on low level features. Those based on fine-grained features represent the similarities of classes on abstract features.



In order to exploit dark knowledge from multi-grained features, we can calculate the weighted average of probability distributions that middle blocks output to get an ensemble distribution that carries multi-views information as Equ \ref{eq:distribution} shows.
 \begin{align}
    D_{E} = \sum^N_iw_i*D_i
	\label{eq:distribution}
\end{align}
 where $D_E$ denotes the ensemble distribution, $w_i$ denotes the weight for the distribution that the ith Extra layer outputs, $N$ denotes the number of the blocks, and the $D_i$ is the distribution that the ith extra layer outputs. In our experiment, the $w_i$ is fixed as $\frac{1}{N}$.
\subsection{Training stage}
In this subsection, we introduce the training stage of LFMA. There are three loss sources in our method:
\begin{itemize}
\item Loss source 1: The Kullback–Leibler divergence(KLD) loss is computed from the ensemble probability distribution to the one from the output of the backbone as Equ \ref{eq:KLD} shows. Here $n$ denotes the number of classes in the dataset, $i$ denotes the ith class, $P(i)$ represents the probability of the ith class in the probability distribution that the backbone outputs and $Q(i)$ denotes such probability in the ensemble probability distribution. The smaller $Loss_{KLD}$ is, the more similar the distributions $P$ and $Q$ are to each other.
\begin{align}
    Loss_{KLD}(P||Q) = -\sum_i^n P(i)\ln{\frac{Q(i)}{P(i)}}
    \label{eq:KLD}
\end{align}
In order to soft the probability distribution over classes, we introduce the temperature $T$ \cite{hinton2015distilling} as Equ \ref{eq:soft_label} shows.
\begin{equation}
    P(i) =\frac{\exp(p_i/T)}{\sum_j\exp(p_j/T)}
    \label{eq:soft_label}
\end{equation}
where $i$ denotes the ith class, $p_i$ is the output of fully connected layers for the ith class, $P(i)$ denotes the probability that the input image belongs to the ith class, and $T$ is the temperature coefficient. A larger $T$ brings a softer distribution.
\item Loss source 2: The Kullback–Leibler divergence(KLD) loss is computed from the ensemble probability distribution to the ones from the output of the extra layers.
\item Loss source 3: Cross entropy(CE) loss under labels of the dataset. The cross-entropy loss is computed using the softmax output of the backbone with labels from the dataset. As Equ \ref{eq:CE_loss} shows, where the $i$ denotes the ith class. 
\begin{equation}
    Loss_{CE}(P, Q)=-\sum_i^n P(i)\log Q(i)
    \label{eq:CE_loss}
\end{equation}
\item Loss source 4: Cross entropy loss under labels of the dataset. The cross-entropy loss is computed using the softmax output of each extra layer with labels from the dataset.

 Here, we take $\alpha$, $\beta$, $\gamma$, $\beta$ to balance four losses as the Equ \ref{eq:loss_3}.
 \begin{equation}
     Loss = \alpha Loss_{1} + \beta Loss_{2} + \gamma Loss_{3} + \delta Loss_{4}
     \label{eq:loss_3}
 \end{equation}
 where $Loss_1$ denotes the Loss source 1, $Loss_2$ denotes the Loss source 2, $Loss_3$ denotes the Loss source 3, $Loss_4$ denotes the Loss source 4.
\end{itemize}

\subsection{Feature Map Cutout}
In order to low memory requirements and improve the efficiency of data augmentation, we put forward a novel data augmentation based on Feature Maps, named Feature Map Cutout(FMC). FMC is to erase random pixies on feature maps in order to improve the general of model since the key position is erased instead of low relevant positions, and achieve the goal of one image corresponding to multiplied feature maps with different augmentations, which can avoid not necessary memory waste because we only need to pass into models one image rather than multiple ones with various augmentations.
We formulate the method as:
\begin{equation}
    M^*_i = Mask_i \odot M_i
    \label{eq: aug}
\end{equation}
where $Mask_i$ denotes a 0-1 matrix with the same size of $M_i$, $M_i$ denotes a feature maps generated by the ith block, and the $M_i^*$ is the feature map after applying element-wise multiplication between $M_i$ and $Mask_i$, which will be passed into the next block. 

\begin{figure}[htb]
    \centering
    \includegraphics[width=1.0\linewidth]{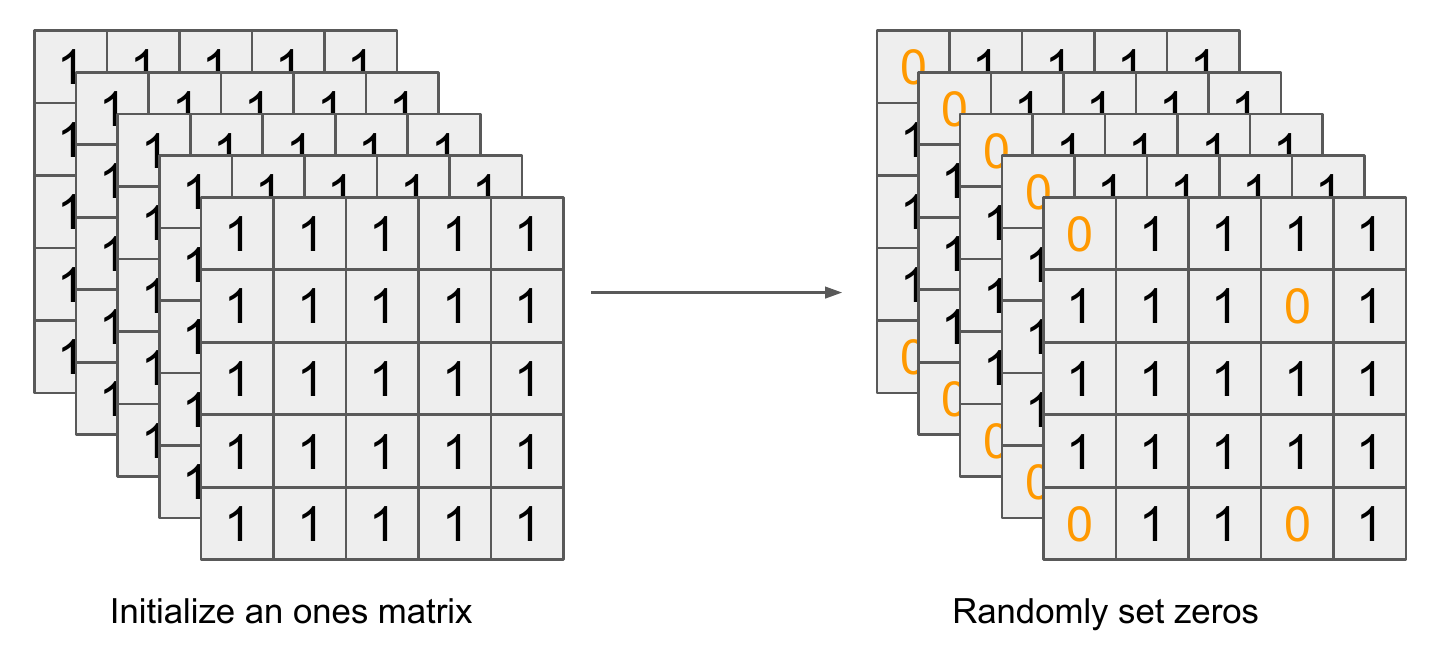}
    \caption{Randomly generate a 0-1 mask matrix}
    \label{fig:Generate_Mask}    
\end{figure}

In particular, $Mask$  is generated randomly and can be formulated as:
\begin{align} 
    Mask_{:,:,:} &= 1
    \label{equ: init}
\end{align}
The mask matrix is initialized to all ones as Equ \ref{equ: init}.
\begin{align} 
    Mask_{:,i,j} &= 0
    \label{equ: one}
\end{align}
Therefore, we randomly choose $i, j$ positions and set all channel of those pixels to 0 as Equ \ref{equ: one} shows. Fig \ref{fig:Generate_Mask} demonstrates how does an mask generated .
\section{Experiments} 
We evaluate our method, MFD, on a varies of network structures(ResNet, WideResNet), datasets(CIFAR-100, Caltech-UCSD Bird(CUB200), Stanford Dogs(Dogs)), which proves our method effective and general. 
\subsection{Datasets}
We demonstrate the effectiveness of MFD on various image classification datasets: CIFAR-100, CUB200 and Dogs. The CIFAR-100 dataset consists of 60000 32*32 colour images in 100 classes, with 600 images per class. CUB200 contains 11788 images of 200 bird species, and each class has 500 images, and the Dogs dataset contains 20580 images of 120 dog species.

\linespread{1.5}
\begin{table}
\setlength{\tabcolsep}{0.6mm}{
\begin{tabular}{|c|c|c|c|c|}
\hline
         & CIFAR100 & CUB200  & Stanford40 &ImageNet\\ \hline
Baseline & 73.08\%  & 51.72\% & 42.97\% &\underline{69.75}\%   \\ \hline
DDGSD    & \underline{76.11}\%  & 57.49\% & 45.21\% &-   \\ \hline
ONE      & 75.41\%  & 54.30\% & 45.05\%  &-  \\ \hline
BYOT     & 76.04\%  & \underline{58.10}\% & \underline{48.00}\%  &-  \\ \hline
SAD      & 74.35\%  & 52.76\% & 43.52\%  &-  \\ \hline
LFMA    & \textbf{79.71}\%  & \textbf{59.65}\% & \textbf{49.21}\%  &\textbf{70.84}\%  \\ \hline
\end{tabular}
}
\caption{Performance comparison on various datasets. The best result is indicated as boldface. The best performance for datasets is indicated as boldface and the second one is indicated as underline.}
\label{table:CIFAR_AND_TINYIMAGENET_PERFORMANCE}
\end{table}


\subsection{Implementations details}
We demonstrate our method on ResNet18, ResNet34, ResNet50, ResNet101 and ResNet110\cite{he2016deep}. To adapt ResNet to small-sized images in CIFAR100, we modify the first convolution layer of them as a kernel size of 3*3, the stride of 1 and the padding of 1. 

We use stochastic gradient descents(SGD) with a momentum of 0.9, an initial rate of 0.1, weight decay of 0.0001. We divide the learning rate by 10 at epoch 100 and 150. The total epoch is set as 200. We use some standard data augmentation methods for image classification task, i.e. random cropping and flipping. For more details, you can access our project on Github. And we demonstrate our methods on 12 NVIDIA Tesla V100.

\subsection{Performance Comparison}
We provide one standard classifier and eight methods of self-distillation as baselines. The standard classifier doesn't utilize the distillation technique and has the same backbone as our approach does. The details of other self-distillation baselines are listed below.
\begin{itemize}
\item \textbf{Baseline} is the standard ResNet18 without self-knowledge distillation methods applied.

\item \textbf{DDGSD}\cite{xu2019data} produce the same output for images that are the same instance yet different data augmentations.

\item \textbf{ONE}\cite{lan2018knowledge} provides an effective training method where the outputs of student networks are aligned to these of teacher one.

\item \textbf{BYOT}\cite{zhang2019your} applies auxiliary network into the backbone and trains them with the ground truth labels from the dataset and the feature map from deeper layers.

\item \textbf{SAD}\cite{hou2019learning} achieves good performance on lane detection with a layer-wise attention self-distillation.

\end{itemize}

\subsubsection{Performance Comparison}
Table \ref{table:CIFAR_AND_TINYIMAGENET_PERFORMANCE} shows the performances of different self-knowledge distillation methods on CIFAR100, CUB200, Dogs and Stanford40 with ResNet18. Compared to the standard classifier, all self-knowledge distillation methods achieve better performances. Furthermore, the proposed method, LFMA, shows better performance than that of other methods. 

\linespread{1.5}
\begin{table}[]
\setlength{\tabcolsep}{0.4mm}{
\begin{tabular}{|c|c|c|c|c|c|p{1cm}p{1cm}p{1cm}p{1cm}p{1cm}p{1cm}p{1cm}|}

\hline
Models                     & Method   & CIFAR100 & CUB200  & Dogs    & Stanford40 \\ \hline
\multirow{3}{*}{Resnet18}  & Baseline & 73.08\%  & 49.91\% & 55.23\% & 41.14\%    \\ \cline{2-6} 
                           & DSN      & 75.99\%  & 53.49\% & 59.88\% & 45.30\%    \\ \cline{2-6} 
                           & LFMA     & 79.71\%  & 60.08\% & 67.80\% & 48.21\%    \\ \hline
\multirow{3}{*}{Resnet34}  & Baseline & 75.74\%  & 48.72\% & 56.10\% & 41.03\%    \\ \cline{2-6} 
                           & DSN      & 76.80\%  & 54.16\% & 60.92\% & 44.32\%    \\ \cline{2-6} 
                           & LFMA     & 79.17\%  & 61.25\% & 69.60\% & 48.69\%    \\ \hline
\end{tabular}}
\caption{Ablation Studies. The Baseline is to train models without self-knowledge distillation techniques applied. The DSN is to train models with deep supervised. The LFMA is to train models with our method.}
\label{table:Ablation_studies}
\end{table}
\subsection{Ablation Study}
To demonstrate the effectiveness of using features with multilevel abstraction to guide network training for performance improvement, we conducted a series of ablation studies.

We apply baseline, deep supervised(DSN) and LFMA on CIFAR100, CUB200, Dogs and Stanford40 datasets with ResNet18, ResNet34, ResNet50 and ResNet101. Table \ref{table:Ablation_studies} shows their classification accuracy. Compared to Baseline, DSN improves the performance on various datasets and network structures. Furthermore, LFMA achieves a more substantial improvement compared with the baseline.



\section{Discussion}
In this section, we visualize attention maps of networks and the features distribution for discussing the reason that LFMA works.

\subsection{Qualitative Attention map comparison}
\begin{figure}[htb]
    \centering
    \includegraphics[width=1.0\linewidth]{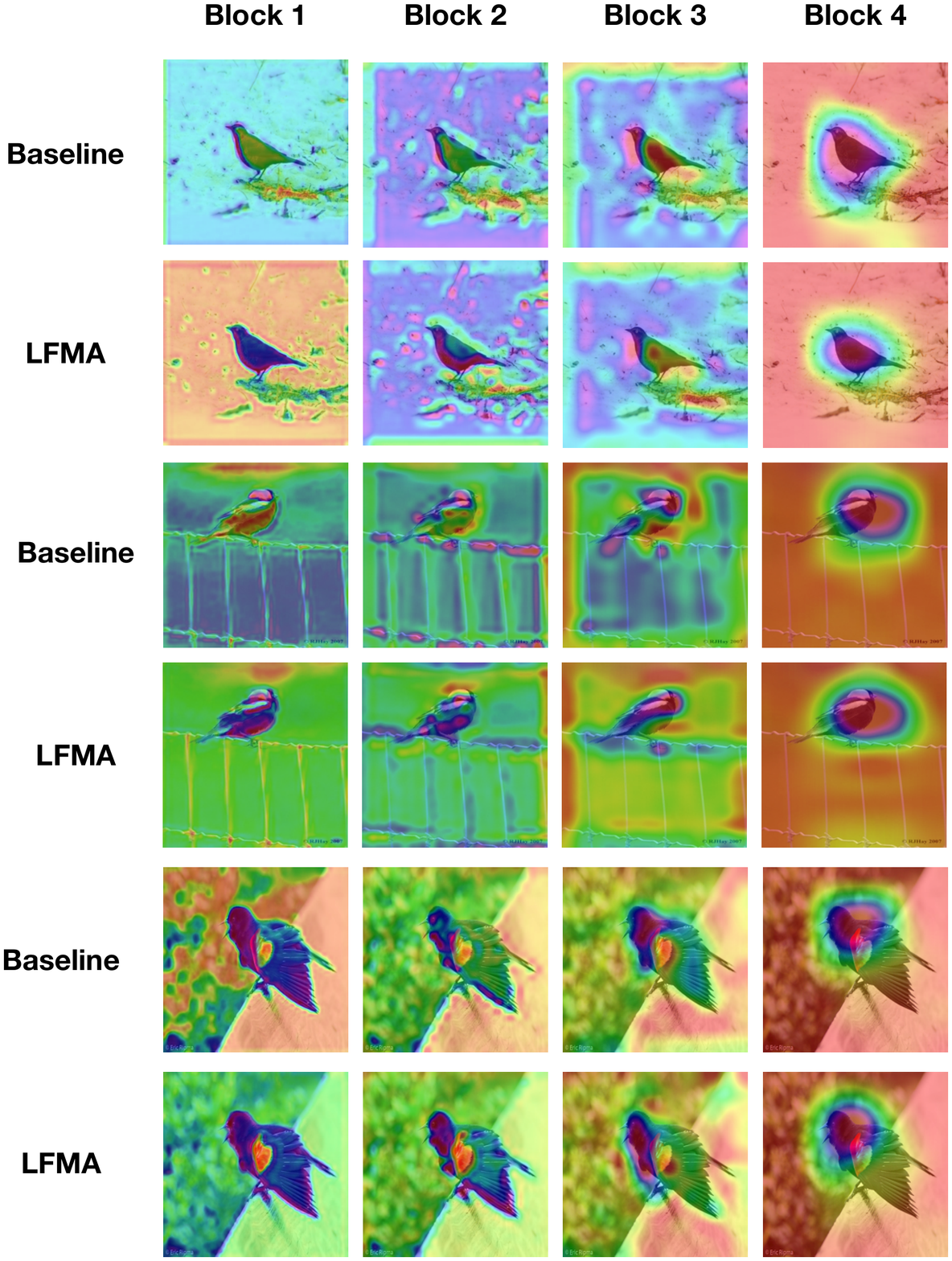}
    \caption{Attention maps comparison between baseline(standard Resnet50) and LFMA(Resnet50 applied LFMA). Each data was taken from CUB200.}
    \label{fig:Attention_Maps}    
\end{figure}

In order to identify that LFMA can exploit features with multilevel abstraction, we conduct qualitiative analysis by visualizing the attention maps of the networks. We train ResNet50 on CUB200 with three different ways: 1. Stanard classifier is that the model is trianed normally; 2. DSN is that the deep supervisied technique is applied for training model; 3. LFMA is that the network is trained with the LFMA tenique. The attention maps from each blocks are visualized as Figure \ref{fig:Attention_Maps} shows.

The network applied with LFMA captures much specific features than the baseline does as the block 1 shows. And LFMA can help network to ignore irrelevant features compared to the baseline as the block 2 shows. The second row of the block 1 shows that the network observe the whole main object but the first one only gets the edge feature. The fourth row of the block 2 pays more attention on object(bird) than the third one does. 
    
\subsection{Features Visualization}
In this subsection, we explore the form of dark knowledge and explain why dose knowledge from middle blocks work for training.

\begin{figure}[htb]
    \centering
    \includegraphics[width=1.0\linewidth]{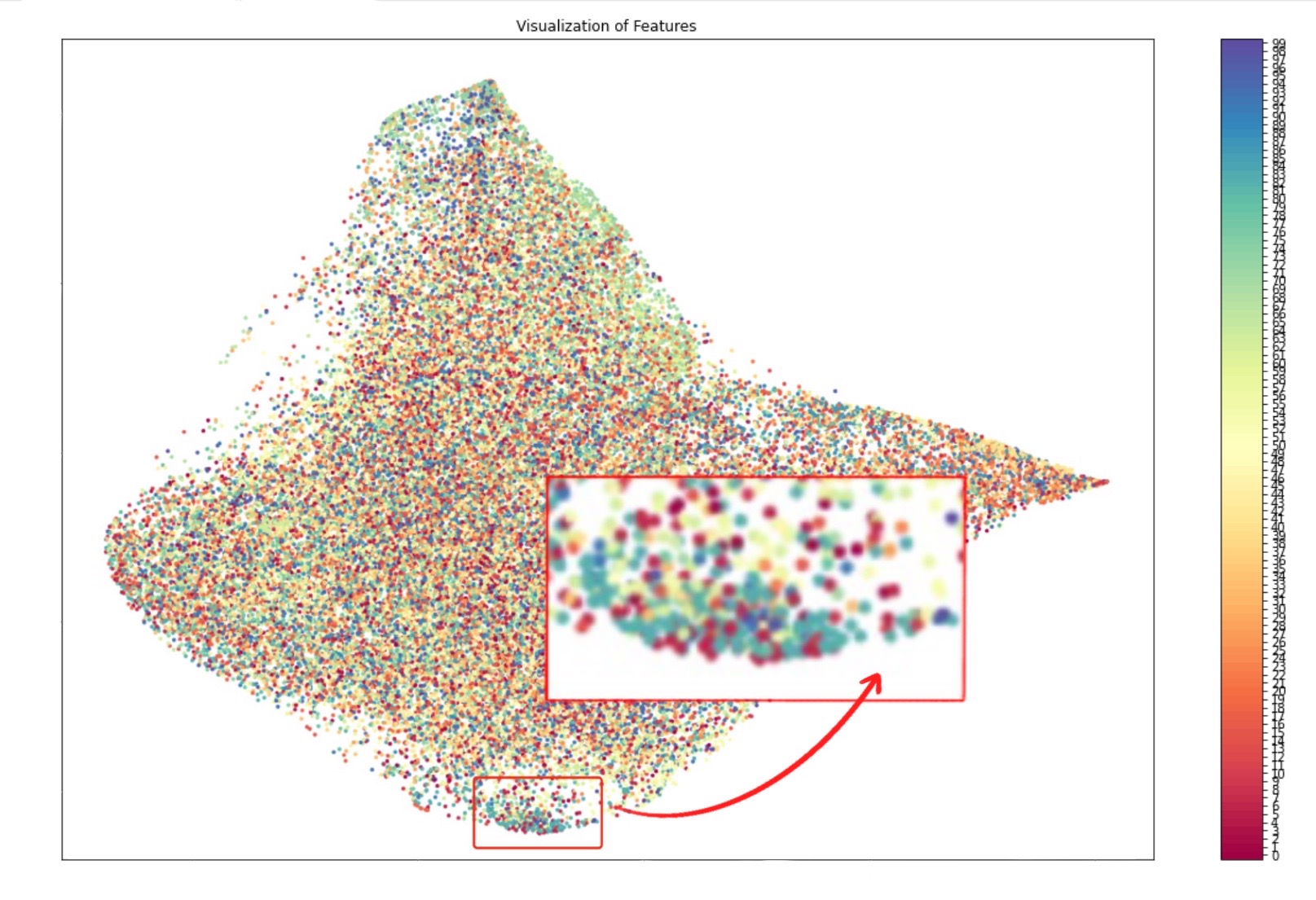}
    \caption{Feature distribution from the first block}
    \label{fig:feature_distribution1}    
\end{figure}

\begin{figure}[htb]
    \centering
    \includegraphics[width=1.0\linewidth]{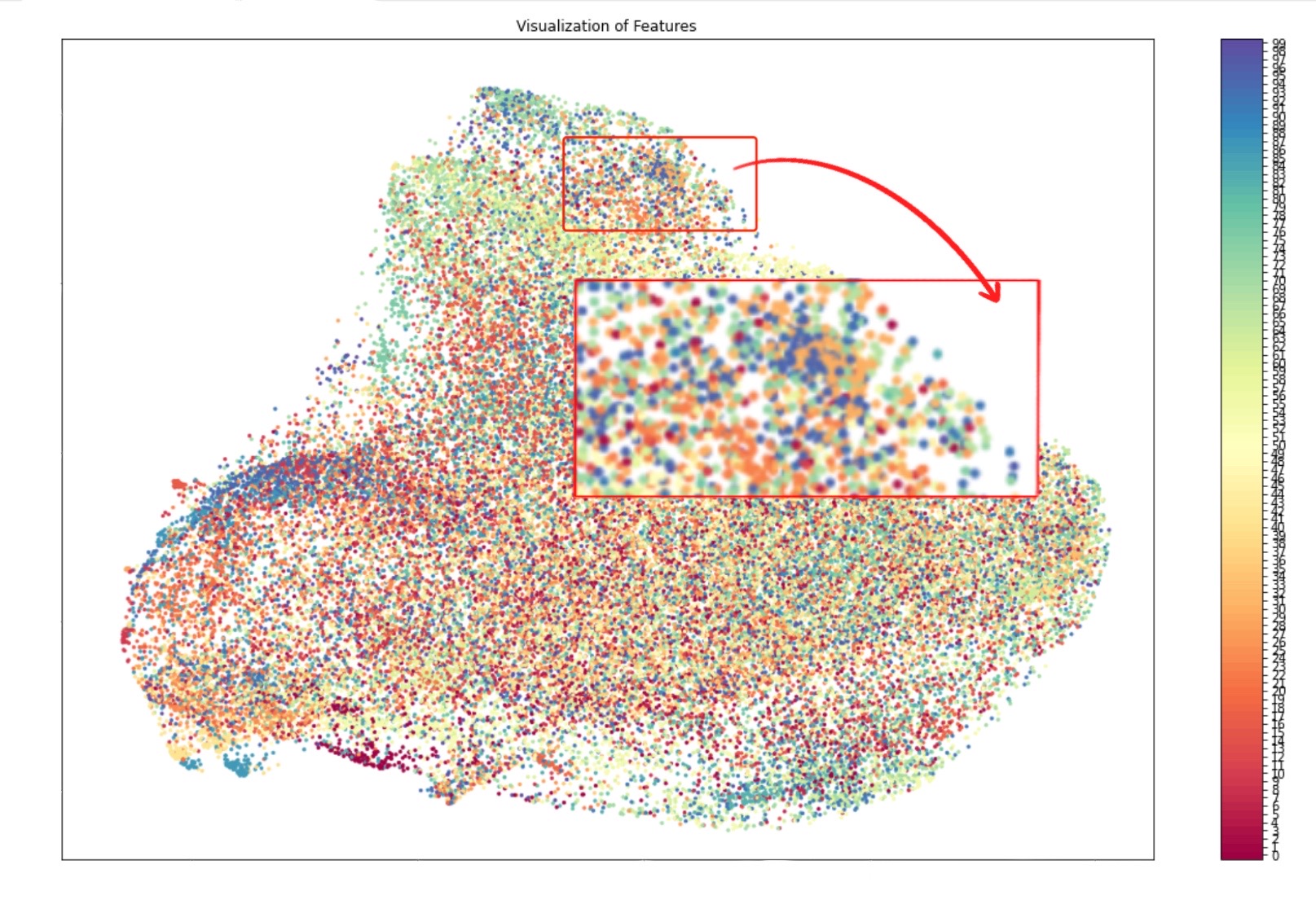}
    \caption{Feature distribution from the second block}
    \label{fig:feature_distribution2}    
\end{figure}

\begin{figure}[htb]
    \centering
    \includegraphics[width=1.0\linewidth]{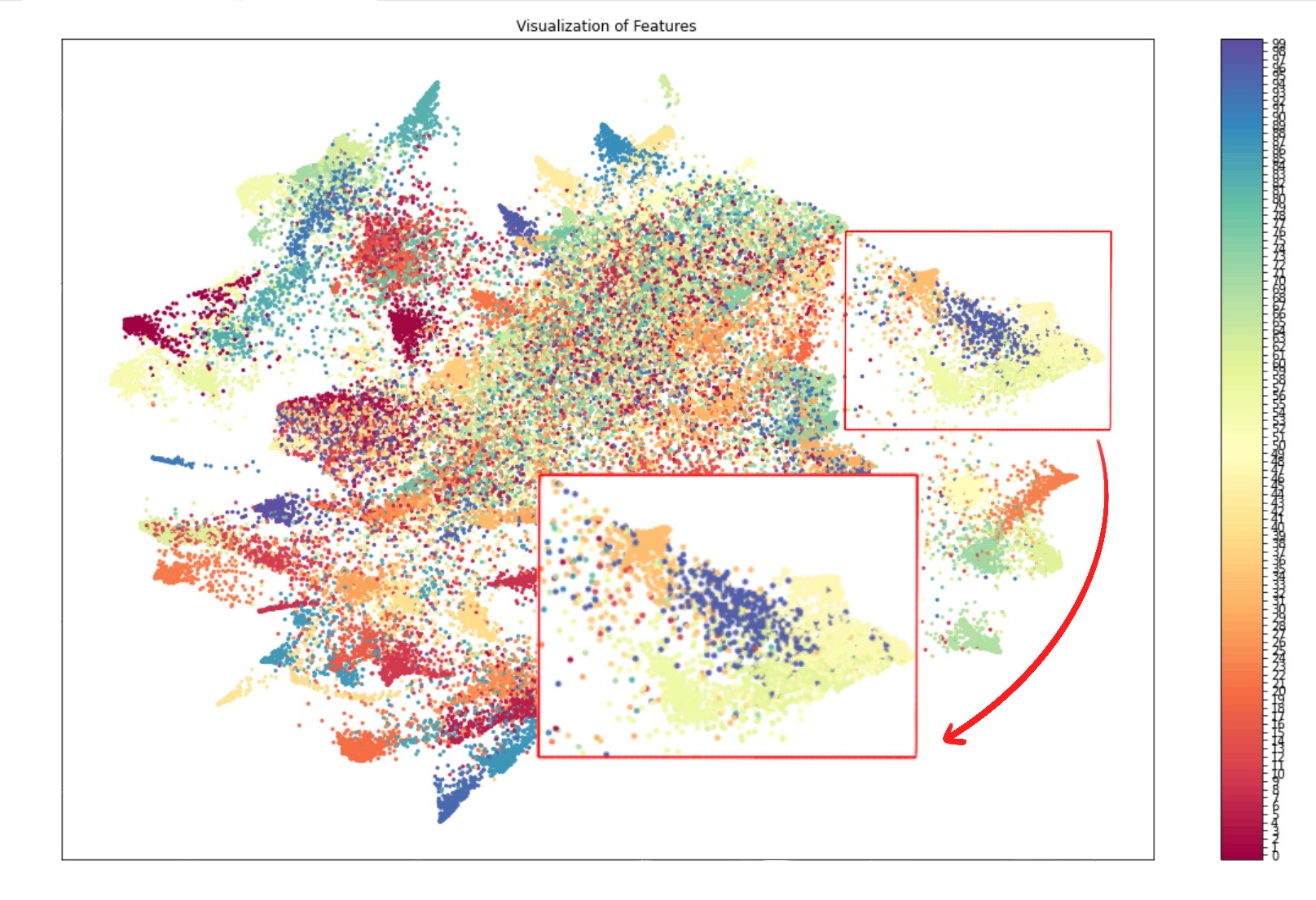}
    \caption{Feature distribution from the third block}
    \label{fig:feature_distribution3}    
\end{figure}

\begin{figure}[htb]
    \centering
    \includegraphics[width=1.0\linewidth]{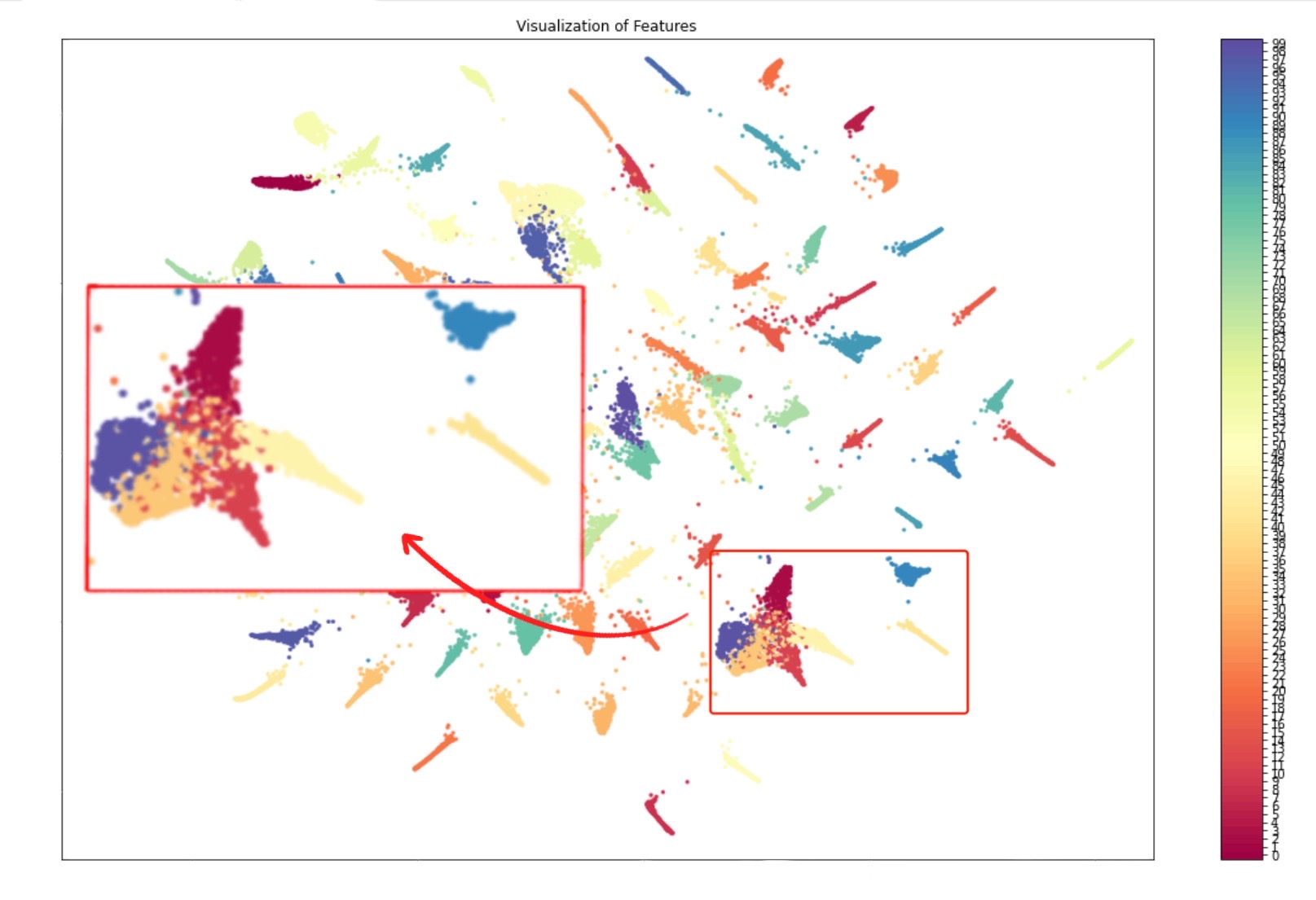}
    \caption{Feature distribution from the forth block}
    \label{fig:feature_distribution4}    
\end{figure}

As the introduction section describes, the main insight of the  paper is that features extracted from different layers can represented different relationships between classes. In order to prove the insight, we visualize the features extracted from various layers as the Figure \ref{fig:feature_distribution1}, \ref{fig:feature_distribution2}, \ref{fig:feature_distribution3} and \ref{fig:feature_distribution4} shows. 

The feature maps extracted from blocks are reduced to 2 dimensions vector with UMAP\cite{mcinnes2018umap}. We visualize them by drawing the first dimension of vectors on vertical, the second dimension of those on the abscissa and labels corresponding input images are colors. All images on the dataset(CIFAR-100) are processed by the network(ResNet18) and their feature maps extracted from various blocks are reduced and visualized. Figure \ref{fig:feature_distribution1} is the representation of features extracted from the block 1, Figure \ref{fig:feature_distribution2} is that of features extracted from the block 2, Figure \ref{fig:feature_distribution3} is that of the features extracted from the block3 and Figure \ref{fig:feature_distribution4} is that of the features extracted from the block4. And images are much similar if the distances of them are closer.

We can observe that the features are much specific and can be easier to classify them correctly as layers deepen. Features with high abstraction extracted from the deep layer can provide useful information for us to classify. Features with low abstraction extracted from the shallow layer cannot provide distinguish information as the deep ones do, but their dark knowledge for self-knowledge distillation. We will analysis details of the features distribution to demonstrate the statement. 

We pay more attentions on the enlarged parts images. The points in Figure \ref{fig:feature_distribution1} are mixed and scattered, while those of the same categories in Figure \ref{fig:feature_distribution3} are clustered. This is why specific features extracted from the deepest layer can be effectively classified.
Vectors reduced by the UMAP technique can be measured similarity with distances\cite{mcinnes2018umap}. The smaller the distance between two vectors, the more similar they are. Some images that don’t share the same categories but they have similar low-level features since they are scattered on a scope as Figure \ref{fig:feature_distribution1} shows. The close distance of different categories that don’t overlap shows their similarity of abstract features as Figure \ref{fig:feature_distribution3} shows. 


These  relationships provided by multilevel abstraction features can help network to get much information from various aspect(shape, edge, and etc.) for achieving a better performance.


\section{Conclusion}
This paper presents a novel way to implement knowledge distillation. It defers from traditional methods where the network is guided by refined knowledge. Our method is to extract knowledge from features with multilevel abstraction. We demonstrate a series of experiments to show it achieves large performance improvements. We believe that the thinking of using multilevel abstraction features can be implemented by other methods such as a large model guided by multi-size models to achieve better performance.

{\small
\bibliographystyle{ieee_fullname}
\bibliography{ref}
}

\end{document}